\documentclass[runningheads]{llncs}
\usepackage{graphicx}
\usepackage{caption}
\usepackage{array, multirow}
\usepackage{subcaption}
\usepackage{bbm}
\usepackage{siunitx}
\usepackage{listings}
\usepackage[flushleft]{threeparttable}
\usepackage[linesnumbered,ruled,vlined]{algorithm2e}
\usepackage[normalem]{ulem}
\useunder{\uline}{\ul}{}
\begin{document}
\title{A Pitfall of Learning from User-generated Data: \\ \textit{In-depth Analysis of Subjective Class Problem}}
%
%
\author{Kei Nemoto\inst{1} \and
Shweta Jain\inst{2}}
\authorrunning{F. Author et al.}
%
\institute{The Graduate Center, CUNY, 365 5th Ave., New York, NY, 10016 USA \\
\email{knemoto@gradcenter.cuny.edu}\and
John Jay College of Criminal Justice, CUNY, 524 West 59th Street, New York, NY, 10019 USA \\
\email{sjain@jjay.cuny.edu}}
\maketitle              
\begin{abstract}
    Research in the supervised learning algorithms field implicitly
    assumes that training data is labeled by domain experts
    or at least semi-professional labelers accessible through
    crowdsourcing services like Amazon Mechanical Turk. With
    the advent of the Internet, data has become abundant and a
    large number of machine learning based systems started being trained
    with user-generated data, using categorical data as true labels. However, little work has been done in the area
    of supervised learning with user-defined labels where users are not necessarily experts and might be motivated to provide incorrect labels 
    in order to improve their own utility from the system. In this article,
    we propose two types of classes in user-defined labels: subjective class and objective class - showing that
    the objective classes are as reliable as if they were provided by domain experts, whereas the subjective classes are subject to bias and manipulation by the user.
    We define this as a \textbf{subjective class issue} and provide a framework for detecting subjective labels in a dataset without querying ‘oracle’. Using this framework, data mining
    practitioners can detect a subjective class at an early stage of their projects,
    and avoid wasting their precious time and resources by dealing with
    subjective class problem with traditional machine learning
    techniques.

\end{abstract}

\section{Introduction} 
Modern supervised learning algorithms share the same goal:
finding a set of parameter values that minimizes a loss function, thus maximizing accuracy, with the error between
predictions made by the learning algorithm and actual labels as feedback. In traditional supervised learning, instances are labeled by domain experts~\cite{Chan98} and trained professionals~\cite{Kubat98}. The high cost of obtaining such labels motivates research in semi-supervised learning \cite{settles.tr09}\cite{Lee13}. 

User generated content  and ratings from  social networks and eCommerce websites are valuable sources of data. Web services such as 
Donorschoose, Kickstarter, and Twitter generously make their datasets available for research and analysis. This data
contains text inputs created by users together with categorical labels selected by users. These new sources of datasets have sparked the 
interests of the data mining community such as \textit{Sentiment Analysis} \cite{Pang05} and \textit{Project Success Prediction} \cite{Greenberg2013CrowdfundingST}. In this
article, we call them ‘user-generated datasets’ and we call the class labels that users provide as user-defined labels. We note users may be inclined to manipulate labels if they perceive a better utility from the system. Thus, user-defined labels show user biases and misinformation.

Researchers have also reported \cite{Tsytsarau11}\cite{Riloff05} that user-generated datasets have relatively lower quality compared to datasets labeled by professionals. Zhu et. al \cite{Zhu04} measure the quality of datasets by the level of attribute noise and class noise. These are crucial to the performance of supervised learning algorithms as they affect the learning of  relationship between attributes and class labels. For example, Riloff et al~\cite{Riloff05} point out that the meaning of a sentence like \textit{``The Parliament exploded into fury against the government...''} is different from when the word \textbf{explosion} appears in an article about war or terrorist attacks. This inconsistency of meanings can be regarded as attribute noise. Class noise, on the other hand, is a problem of having inconsistent instances assigned the same class label or inconsistent labels  assigned to similar instances. While attribute noise has received some attention, class noise in user-generated data has not been discussed in literature. 

In this paper, we seek to answer the following question: \textit{How reliable user-defined class labels are compared to labels created by professionals?} To answer this question, we define two classes in user-defined labels - the objective classes and the subjective classes. As we show in section 4, a classifier that learns from objective classes yields a performance which is comparable to one learned from data labeled by experts which is quite different for classifiers learning from subjective classes. Subjective class problem can be considered as a family of class noise issue, but specific to user-generated datasets. Unlike mislabeling in traditional class noise problem \cite{Smyth96}, label corruption happens when users are motivated to provide certain labels, e.g. claiming low income level to improve their utility from the system or choosing relatively neutral ratings  due to personal sentiments or biases unrelated to the object they are rating. This type of label corruption in subjective class is fundamentally different in a sense that the whole labels are assigned to instances by users and the dataset shows much higher label corruption rate. As we demonstrate in section 3, an example of subjective class is the poverty level of public schools provided on a crowd-funding website, Donorschoose. Of the poverty levels assigned to projects, we found that the error rate is more than 95\%, in contrast to the average error rate of 5\% ~\cite{Redman98}. 

In order to detect subjective class problem, without obtaining true labels, we propose a framework with three quantitative measures in section 4. First, we compare the classification performance of objective and subjective classes within the same dataset.  Second, we visually show the (in)consistency of the two classes using t-SNE plot with Doc2Vec feature. Third, we propose a Normalized Feature Indicative Score (NFIS) to quantify the feature inconsistency in class labels in a subjective class.  Finally in Section 5, we apply our framework to customer rating datasets to show how our framework detects the subjective class problem. The source code used in this study are available online\footnote{\url{https://github.com/box-key/Subjective-Class-Issue}}.  

\section{Related Work}  Previous work \cite{Sheng08}\cite{Zhu04}\cite{Bekker16}\cite{Brodley99} has shown that the performance of classifiers decreases with increase in class noise in the datasets. As Zhu et. al \cite{Zhu04} described, noise can be categorized into attribute noise and class noise, and the impact of attribute noise depends on the correlation between an attribute and the target classes. The authors showed that in a dataset with 50\% corruption rate, the classification performance drops as low as 40\%.

In traditional supervised learning tasks, Brodley and Friedl \cite{Brodley99} summarized major causes of mislabeling as subjectivity, data-entry error, or inadequacy of the information to label instances. Smyth \cite{Smyth96} argued that subjective labeling errors occur in domains where experts naturally disagree on their decisions such as in medical diagnosis. 

Brodley and Friedl \cite{Brodley99} proposed a filtering method to detect mislabeled instances before training a classifier. This method trains a set of different classifiers on the training data with n-fold cross validation as a filtering committee. Then, it performs majority voting or consensus voting to decide which instances are to be used for training. Their empirical results show that ensemble filters performed better than filters with a single classifier.

While filtering techniques are effective in handling class noise issue, Bekker and Goldberger \cite{Bekker16} pointed out that the filtering methods are not scalable and removing instances is sometimes not feasible in a small training set. To overcome this difficulty,   researchers \cite{Bekker16} \cite{Lee13} \cite{Mnih12} have proposed a semi-supervised deep neural network model trained on a dataset containing class noise without removing mislabeled instances. Mihn and Hinton \cite{Mnih12} proposed a modified loss function robust to a specific class noise in their task. Bekker and Goldberger \cite{Bekker16} developed a probabilistic model which incorporates true labels of instances as latent variable into a deep neural network to learn noise distribution.

While filtering techniques are to detect mislabeled examples before training, active learning aims to achieve high classification performance with selected instances \cite{settles.tr09}. \textit{Active Learners} are first trained by a portion of instances from a labeled
training set, then actively query instances that improve classification performance the most from a pool of unlabeled instances. Active learning algorithms vary depending on the minimization objective along with a querying strategy. For example, Uncertainty Sampling chooses instances of which a classifier is most uncertain to decide memberships \cite{Lewis94}. Expected Gradient Length (EGL) queries instances that expect to give the biggest change to the current model \cite{settles.tr09}\cite{NIPS2007_3252}. Settles et al. \cite{NIPS2007_3252} proposed a method that uses the gradient to measure the degree of change. Cohn et al. \cite{Cohn94} introduced an active learning method that selects
instances that minimize the future variance to reduce classification
errors. These approaches may not be applicable to subjective classes since they implicitly require a `reliable’ base classifier that is capable of querying useful instances to be used in training.

It is worth noting that Sheng et al. \cite{Sheng08} addressed the issue
of low quality labels made by less-than-expert outsourcing labeling services such as Rent-A-Coder or Amazon’s Mechanical Turk. They pointed out the importance of quality control over classes made by labelers who don’t possess domain knowledge. In their work, repeated-labeling method by labelers with more than 50\% of labeling accuracy is proven to be effective to maintain the quality of labelsets. 

\section{Demonstration of Subjective Class Problem}
There are some tasks that require user-given categorical values as supervised class, e.g. customer rating prediction \cite{Pang05} or mood classification using micro-blog tags\cite{Mishne05}. In this article, we use a user-generated dataset to demonstrate subjective class problem by comparing user-defined poverty level to poverty level obtained through the census data.
\vspace{-.2in}
\lstdefinelanguage{json}{
    basicstyle=\small,
    numbers=left,
    numberstyle=\scriptsize,
    stepnumber=1,
    numbersep=2pt,
    showstringspaces=false,
    breaklines=true,
    frame=lines,
}
\begin{figure}[h]
\centering
\begin{lstlisting}[language=json,firstnumber=1, label={lst:donorschoose}, caption={the format of project data in Donorschoose dataset}]

{"project": {
  "_projectid": int,
  "school_ncesid": 12 characters,
  "poverty_level": categorical,
  "grade_level": categorical,
  "primary_focus_area": categorical,
  "title": max 30 words,
  "short_description": max 290 words,
  "need_statement": max 660 words,
  "essay": max 1718 words
 }
}
\end{lstlisting}
\end{figure}
\vspace{-.35in}
\subsection{Dataset Schema}
Listing \ref{lst:donorschoose} shows the schema from part of the dataset from Donorschoose.org \footnote{\url{https://research.donorschoose.org/t/download-opendata/33}}, which we use in this work. Donorschoose is a web-based crowd donation platform that allows teachers in public schools to create project pages. Donors visit the project pages and choose to donate to the projects directly. This dataset contains 1,201,597 past projects posted with details of those projects including values in Listing \ref{lst:donorschoose}. The full dataset has 54 columns. Examples of columns we don’t use in this study are the address of schools, funding status (i.e. whether a project completed or expired) and the final amount of donation a project received. We build text classifiers that take as input `title', `short\_description', `need\_statement' and `essay' and outputs predicted labels for \textit{poverty\_level}, \textit{grade\_level} and \textit{primary\_focus\_area} as target classes. Next, we demonstrate that \textit{poverty\_level} has the subjective class problem by obtaining true poverty level of public schools using the census data.

\subsection{Obtaining True Poverty Level}
According to National Center for Education Statistics (NCES), one way to measure the poverty level of a school is to look up the percentage of the students who are eligible for Free or Reduced-Price Lunch program, a.k.a. FRPL  \footnote{\url{https://nces.ed.gov/programs/coe/indicator\_clb.asp}}. NCES defines poverty levels of schools based on the percentage of students who qualify for FRPL (Table~\ref{tab:poverty_level_def}). 
Following this definition, we obtain true poverty level of schools by estimating the percentage of students who fall into the poverty category by dividing the estimated total population of 5 to 17 years old of each school district by the estimated number of relevant children 5 to 17 years old in poverty who are related to the householder in 2017 provided
by Small Area Income and Poverty Estimates (SAIPE) \footnote{\url{https://www.census.gov/data/datasets/2017/demo/saipe/2017-school-districts.html}}, then we map those percentages into poverty levels. Finally we match the above poverty rate and poverty level table to schools in Donorschoose dataset by their school district ids (school\_ncesid in Listing \ref{lst:donorschoose}). 

\begin{table}[h]
    \vspace{-0.3in}
    \centering
    \begin{threeparttable}[width=0.2\textwidth]
      \caption{Poverty Level Definition}
      \label{tab:poverty_level_def}
         \begin{tabular}{c|c}
            \hline
            \  Estimated Poverty Rate & Poverty Level \\
            \hline
            \ 0-25\%  & Low Poverty\\
            \ 25\% - 50\% & Moderate Poverty\\
            \ 50\% - 75\% & High Poverty\\
            \ 75\% - 100\% &  Highest Poverty\\
          \hline
        \end{tabular}%
    \end{threeparttable}
    \vspace{-0.4in}
\end{table}

\subsection{True poverty level vs User-given poverty level}
\begin{figure}[h]
  \vspace{-0.3in}
  \centering
  \begin{subfigure}{0.42\textwidth}
    \includegraphics[width=\textwidth]{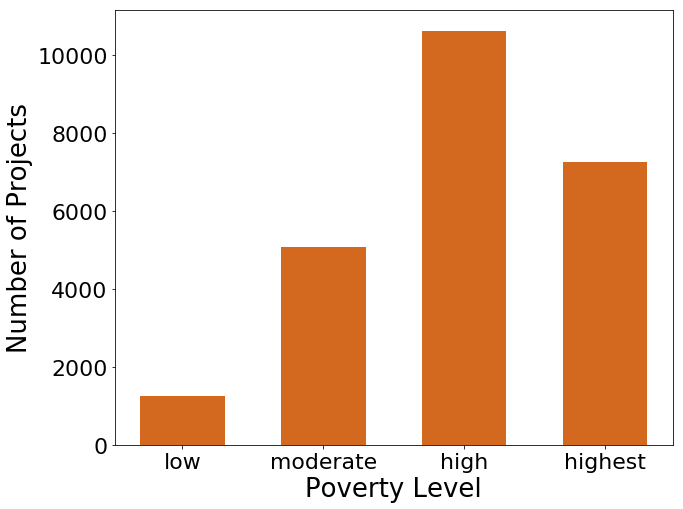}
    \caption{User-defined Poverty Level}
    \label{fig:user_given_poverty_level}
  \end{subfigure}
  \hfill
  \begin{subfigure}{0.48\textwidth}
    \includegraphics[width=\textwidth]{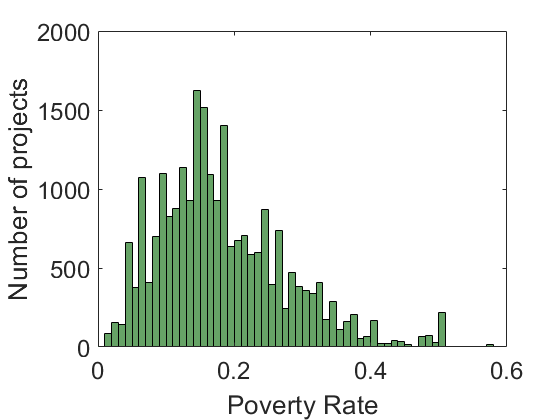}
  \caption{Poverty level  from census data}
  \label{fig:true_poverty_rate}
  \end{subfigure}
  \caption{The distribution of poverty levels given by users and census data}
    \vspace{-0.2in}

\end{figure}

To make the argument concrete, we only used schools located in the school districts with no more than two schools in Donorschoose dataset for plotting the distribution of true poverty rate (Figure \ref{fig:true_poverty_rate}). In Figure \ref{fig:user_given_poverty_level}, we plot the user given poverty labels in projects posted from the same schools. As Figure \ref{fig:user_given_poverty_level} shows, most   users chose `highest poverty' or `high poverty' for the poverty level of their classrooms. However, Figure \ref{fig:true_poverty_rate} shows that there are few schools belonging to high poverty or highest poverty in our definition, and most schools are either moderate poverty level or low poverty level.

\begin{figure}[h]
  \vspace{-0.3in}

  \centering
  \includegraphics[width=0.7\textwidth]{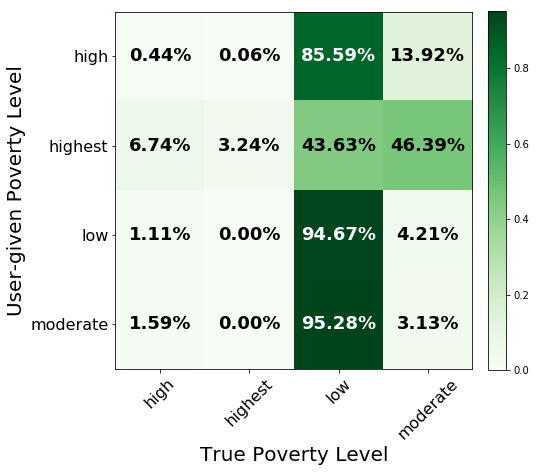}
  \caption{Matrix of user-given poverty level vs true poverty level.}
  \label{fig:poverty_level_matrix}
  \vspace{-0.5in}
\end{figure}

Figure \ref{fig:poverty_level_matrix} shows the poverty levels given by users don't reflect their actual poverty levels in our definition (Table \ref{tab:poverty_level_def}). An element at row ``highest poverty'' and column ``highest poverty'' in Figure \ref{fig:poverty_level_matrix} shows that only 3.24\%  of the projects assigned ``highest poverty level'' by user actually belong to ``highest poverty level'' and more than 90\% of them belong to ``low poverty level'' or ``moderate poverty level''. An element at row `'low poverty'' and column ``low poverty'' tells us that 95\% of users who chose ``low poverty level'' belong to that label. However, we can infer, from ``low poverty'' column, that most of users who labels other than '`low poverty level'' actually belong to ``low poverty level''.

It is not hard to assume that there is an incentive to choose higher poverty levels. This leads us to the conclusion that training a supervised algorithm with a subjective class would end up generating a poor classifier. Moreover, unlike class noise problem
proposed in previous work \cite{Sheng08}\cite{Zhu04}\cite{Bekker16}\cite{Brodley99}, the mislabeling in a subjective class have a more coherent pattern. Though class noise techniques proposed in previous work are required to build a base classifier to detect class noise, it’s hard to do so with subjective class.  Therefore, data mining practitioners need to consider whether their supervised classes are subjective or not during their projects. In the next section, we show our analysis of the difference between subjective class and objective class to propose a guideline for finding a subjective class without using an oracle.

\section{Analysis of Subjective Class and Objective Class}
In this section, we address two questions: ‘is objective class as reliable as classes defined by domain experts?’ and ‘how to detect whether user-defined class has a subjective class problem without using true labels?’. To answer these questions, we compare objective class and
subjective class by the following three criteria:

\begin{itemize}
    \item Classification Performance.
    \item Class (In)consistency by plotting data points by Doc2Vec and t-SNE method.
    \item Feature (In)consistency by Normalized-Feature Indicative Score (section 4.3).
\end{itemize}

\subsection{Classification Performance}

We trained text classifiers to predict the \textit{grade-level}, \textit{primary subject area}, and \textit{poverty-level} of each project separately  in Donorschoose dataset based on their project descriptions. The purpose is to compare how classifiers perform when the data consists of labels in the subjective class compared to when the labels are in the objective class.

\begin{table}[h]
\small
    \scalebox{0.7}{
    \begin{subtable}[t]{0.45\linewidth}
        \begin{tabular}{|c|c|}
            \hline
            \ Label Name & Frequency \\
            \hline
            \ Highest Poverty & 94,950 \quad 59.13 (\%)\\
            \ High Poverty & 39,117 \quad 24.36 (\%)\\
            \ Moderate Poverty & 22,187 \quad 13.82 (\%)\\
            \ Low Poverty & 4,305  \quad 2.68 (\%)\\
            \ - & -\\
            \ - & -\\
            \ - & -\\
            \hline
        \end{tabular}
        \caption{Poverty Level}
        \label{fig:dist_poverty_level}
    \end{subtable}
    }
    \scalebox{0.7}{
    \begin{subtable}[t]{0.42\linewidth}
        \begin{tabular}{|c|c|}
                \hline
                \ Label Name & Frequency \\
                \hline
                \ Grades PreK-2 & 51,526 \quad 32.1 (\%)\\
                \ Grades 3-5 & 49,868 \quad 31.06 (\%)\\
                \ Grades 6-8 & 30,536 \quad 19.02 (\%)\\
                \ Grades 9-12 &  28,629 \quad 17.61 (\%)\\
                \ - & -\\
                \ - & -\\
                \ - & -\\
                \hline
        \end{tabular}
        \caption{Grade Level}
        \label{fig:dist_grade_level}
    \end{subtable}
    }
    \scalebox{0.7}{
    \begin{subtable}[t]{0.42\linewidth}
        \begin{tabular}{|c|c|}
                \hline
                \ Label Name & Frequency \\
                \hline
                \ Literacy \& Language & 64,351 \quad 40.08 (\%)\\
                \ Math \& Science & 45,145 \quad 28.12 (\%)\\
                \ Music \& Arts & 14,374 \quad 9 (\%)\\
                \ Special Needs & 11,646  \quad 7.3 (\%)\\
                \ Applied Learning & 10,537  \quad 6.6 (\%)\\
                \ History \& Civics & 7,812  \quad 4.9 (\%)\\
                \ Health \& Sports & 6,694  \quad 4.2 (\%)\\
                \hline
        \end{tabular}
        \caption{Primary Focus Area}
        \label{fig:dist_subject}
    \end{subtable}
    }
    \caption{The distribution of labels in target classes}
    \label{tab:class_distribution}
\end{table}
\vspace{-.2in}
\textbf{Experiment Settings:}
 During the training, we only use instances containing between 2,500 and 4,000 characters. Within the instances,
101,836 (70\%) corpora are used in the training set and 39,318 (30\%) corpora are used in the test set. Table \ref{tab:class_distribution} shows the class distribution of each class in the training set.

 On the training algorithm, we used a linear Support Vector Machine (SVM) as it is reported \cite{Yang99} to yield the best performance
on text classification tasks. With a 10-fold cross validation, a linear SVM with the penalty rate of 5 shows the best performance on average over the three target classes. We choose Tf-Idf Bigram BOW (Bag of Words) feature to vectorize documents in the dataset, which results in 7,252,774 unique bigram terms in the vocabulary. As a performance
measure, we used f1-macro score (equation \ref{eqn:f1score} and \ref{eqn:f1macroscore}); the average over f1-score of each class label.

    \begin{equation}
        F_{1}(c) = 2*\frac{precision(c)*recall(c)}{precision(c) + recall(c)}
        \label{eqn:f1score}
    \end{equation}
    
    \begin{equation}
        F_{1 macro} = \frac{1}{M}\sum_{c=1}^{M} F_{1}(c)
        \label{eqn:f1macroscore}
    \end{equation}
    
, where M is the number of class labels and c indicaes a class label:\\ $c \in \{1 \dots M\}$.

\textbf{Results:}
Table \ref{tab:classification_result} shows the recall and precision of each class label with the f1-macro score of a class in the last row. The results indicate that a learning algorithm produced classifiers favoring the majority class labels in poverty level and primary focus area class due to its highly skewed class distribution \cite{Provost_machinelearning}\cite{Weiss01theeffect}. This is shown by the high precision and considerably low recall on minority classes. However, the recall of minority classes in primary focus area, e.g. `Applied Learning’, `History \& Civics’, and `Health \& Sports’ show much higher recall than the ones of `Low Poverty’ in Poverty Level, in spite of higher class skew and more class elements in primary focus area. In addition, f1-macro scores for grade level and primary focus area indicate  learning from objective classes would yield promising results. 

On the other hand, a classifier learned with a subjective class yields significantly lower f1-macro score. This imbalanced f1-macro scores among different classes within the same dataset can be used as an useful clue to detect a subjective class in a dataset.

\begin{table}[h]
\small
    \scalebox{0.6}{
    \begin{subtable}[t]{0.52\linewidth}
        \begin{tabular}{|c|c|c|c|}
        \hline
         \ Label Name & Recall & Precision & F1 score \\
                \hline
                \ Highest Poverty & 0.89 & 0.75 & 0.81\\
                \ High Poverty & 0.43 & 0.56 & 0.49\\
                \ Moderate Poverty & 0.42 & 0.6 & 0.5\\
                \ Low Poverty & 0.23 & 0.88 & 0.36\\
                \ - & - & - & -\\
                \ - & - & - & -\\
                \ - & - & - & -\\
                
        \hline
        \multicolumn{2}{|c|}{F1-macro score} & \multicolumn{2}{|c|}{\textbf{0.54}} \\
        \hline
        \end{tabular}
        \caption{Poverty Level}
    \end{subtable}
    }
    \scalebox{0.6}{
    \begin{subtable}[t]{0.5\linewidth}
        \begin{tabular}{|c|c|c|c|}
         \hline
        \ Label Name & Recall & Precision & F1 score \\
                \hline
                \ Grades PreK-2 & 0.88 & 0.89 & 0.89\\
                \ Grades 3-5 & 0.83 & 0.79 & 0.81\\
                \ Grades 6-8 & 0.73 & 0.82 & 0.77\\
                \ Grades 9-12 & 0.89 & 0.87 & 0.88\\
                \ - & - & - & -\\
                \ - & - & - & -\\
                \ - & - & - & -\\
        \hline
        \multicolumn{2}{|c|}{F1-macro score} & \multicolumn{2}{|c|}{0.84} \\
        \hline
        \end{tabular}
        \caption{Grade Level}
    \end{subtable}
    }
    \scalebox{0.6}{
    \begin{subtable}[t]{0.5\linewidth}
        \begin{tabular}{|c|c|c|c|}
        \hline
        \ Label Name & Recall & Precision & F1 score \\
                \hline
                \ Literacy \& Language & 0.88 & 0.8 & 0.81\\
                \ Math \& Science & 0.83 & 0.82 & 0.72\\
                \ Music \& Arts & 0.86 & 0.88 & 0.87\\
                \ Special Needs & 0.67 & 0.73 & 0.7\\
                \ Applied Learning & 0.44 & 0.61 & 0.51\\
                \ History \& Civics & 0.65 & 0.8 & 0.72\\
                \ Health \& Sports & 0.81 & 0.8 & 0.81\\
                
        \hline
        \multicolumn{2}{|c|}{F1-macro score} & \multicolumn{2}{|c|}{0.76} \\
        \hline
        \end{tabular}
        \caption{Primary Focus Area}
    \end{subtable}
    }
    \vspace{.1in}
    \caption{Recall, Precision, F1 score for each label in each class with its f1-macro score}
    \label{tab:classification_result}
\end{table}
\vspace{-.6in}
\subsection{Class Consistency}
\begin{figure*}[h]
  \centering
  \begin{subfigure}[b]{0.48\textwidth}
    \caption{Grade Level}
    \includegraphics[width=\textwidth]{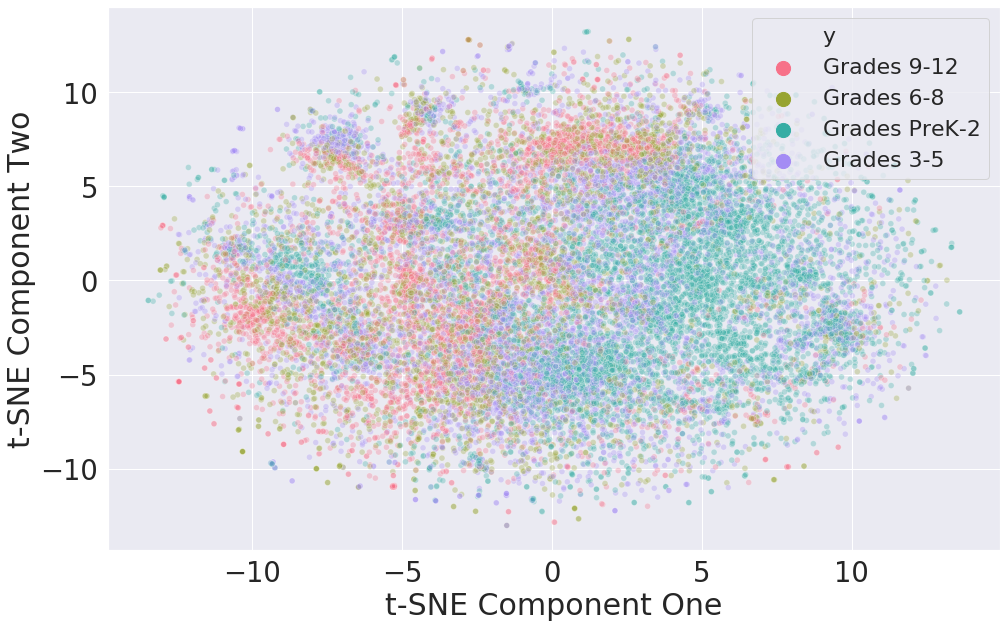}
    \label{fig:f2}
  \end{subfigure}
  \hfill
  \begin{subfigure}[b]{0.48\textwidth}
    \caption{Poverty Level}
    \includegraphics[width=\textwidth]{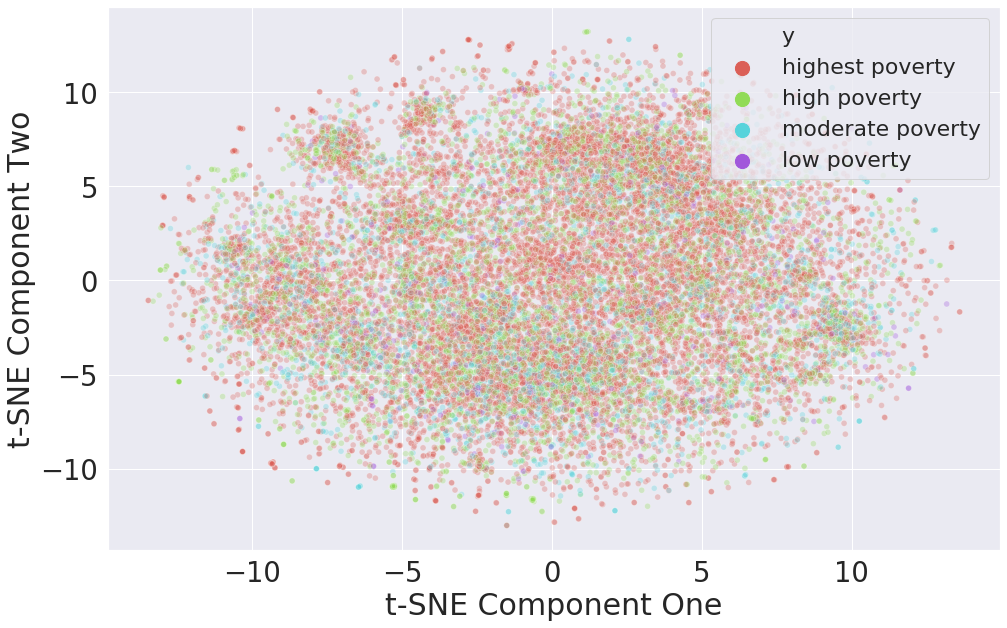}
    \label{fig:f3}
  \end{subfigure}
  \hfill
  \begin{subfigure}[b]{0.6\textwidth}
    \caption{Primary Focus Area}
    \includegraphics[width=\textwidth]{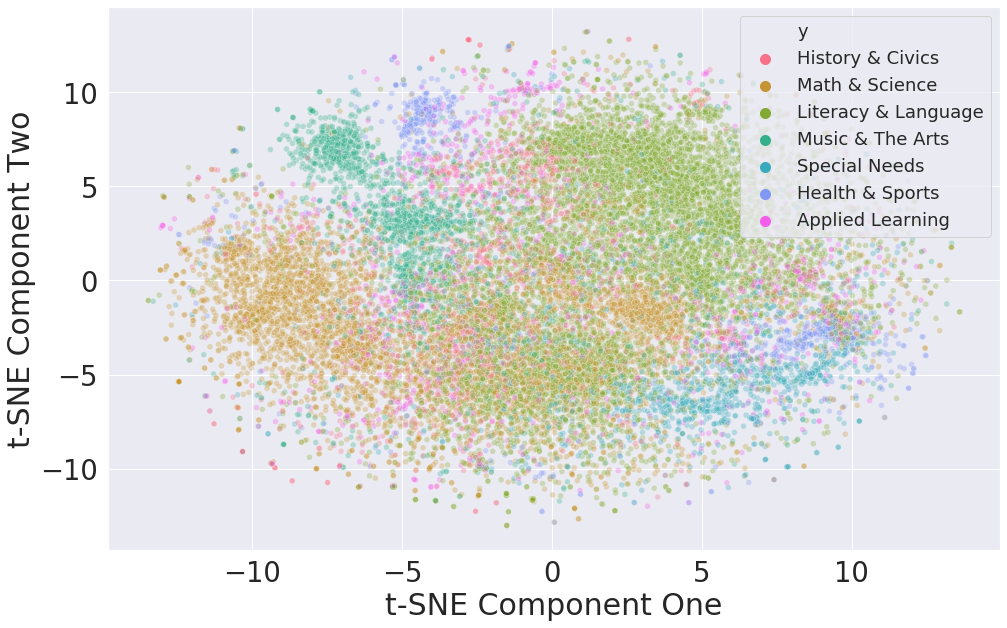}
    \label{fig:f1}
  \end{subfigure}  
  \vspace{-.07in}
  \caption{Data points for each class by t-SNE}
  \label{fig:tsne}
  \vspace{-.07in}
\end{figure*}

Results from the previous sections lead us to a hypothesis that class elements in subjective class are inseparable, or inconsistent, whereas ones in objective class are separable (consistent). We test this hypothesis by   compressing our corpora into two dimensional space with Doc2Vec
feature and t-SNE method \cite{vanDerMaaten2008} and plotting the results in Figure~\ref{fig:tsne}.

\textbf{Doc2Vec:}
Doc2Vec~\cite{Mikolov13b} is an extension of Word2Vec \cite{Mikolov13a}, which maps documents into n feature vector space using a trained neural probabilistic language model \cite{Bengio2003}. Mikolov et al~ \cite{Mikolov13b} proposed Distributed Memory version of Paragraph Vector (PV-DM), a model that learns parameters by adding extra layer called Paragraph ID to the Continuous Bag of Words (CBOW) model for Word2Vec. On the other hand, Distrbuted Bag of Words version of Paragraph Vector (PV-DBOW) learns parameters by predicting context words by paragraph ID layer similar to a skip-gram model in Word2Vec. As previous work \cite{lau-baldwin-2016-empirical}\cite{Mikolov13b} shows, Doc2Vec outperformed other document representation methods such as n-gram models. In our experiment, we trained PV-DM model and PV-DBOW model with 1,201,597 corpora in Donorschoose dataset by vector size = 300, window = 8, minimum word count = 10, iteration = 10 for each model.

The performance of Doc2Vec model depends on a task \cite{lau-baldwin-2016-empirical}. Thus, to decide which model to use as a feature, we trained a SVM classifier for poverty level, grade level and primary focus area and compute f1-macro score for each model. In addition, as suggested in \cite{Mikolov13a}, we tested a combined PV-DM model and PV-DBOW model since they have been found to yield the best performance. 

\begin{table}[h]
    \centering
    \begin{threeparttable}[width=0.3\textwidth]
         \begin{tabular}{|c|c|c|c|}
            \hline
            \ Class & DM & DBOW & DM + DBOW\\
            \hline
            \ Grade Level  & 0.61 & 0.69 & 0.7\\
            \ Poverty Level & 0.24 & 0.31 & 0.33\\
            \ Primary Subject Area & 0.67 & 0.69 & 0.69\\ 
            \hline 
        \end{tabular}
    \vspace{0.1in}
     \caption{The f1-macro scores of classifiers trained with differnt Doc2Vec models.}
     \end{threeparttable}
     \label{tab:doc2vec}
\vspace{-.4in}
\end{table}

As in Table 4, our DBOW model showed higher
performance than our DM model for poverty level and grade level, and it shows almost the same score for Primary Focus Area. Furthermore, our combined model doesn’t improve the performance significantly. Thus, to keep the dimension low, we decided to use our DBOW model in this experiment.

\textbf{t-SNE dimension reduction:}
Maaten and Hinton \cite{vanDerMaaten2008} proposed t-Distributed Stochastic Neighbor Embedding (t-SNE) which is used to reduce the dimensionality of  datasets   while preserving pairwise distance between points close to each other in the high dimensional space. The algorithm attempts to minimize the KL divergence between the probability distribution of data points in high dimensional   and  low dimensional spaces by gradient descent during the training. In this manner, this method successfully preserves the actual distance between similar data points.

We trained our model with randomly chosen 20,000 corpora mapped into Doc2Vec vectors from our dataset by learning rate = 1000, perplexity = 145, iteration = 3000. From Figure \ref{fig:tsne}, we can find several meaningful clusters in grade level (b) and primary focus area (c). However, as we hypothesized, instances in Poverty Level (a) are mixed-up, and Figure \ref{fig:tsne} doesn't show any separable clusters. This observation confirms that a subjective class is more inconsistent than objective class and t-SNE plot with Doc2Vec features can be used to check class consistency of a dataset.

\subsection{Feature Inconsistency in Majority Class}

From section 3, we confirm that roughly 95\% of users who chose ‘low poverty level’ actually belong to ‘low poverty level’ in our definition, while more than 90\% of users who chose other poverty levels would belong to lower poverty levels. This observation leads to an assumption that there are more indicative features (n-gram terms in our experiment)
to low poverty level than to other classes. Intuitively, project organizers who arbitrarily chose labels may not include terms that can be a clue to the actual poverty level of their school, whereas ones who belong to low poverty school would include words indicative to their poverty level. In this experiment, we propose a method called Normalized Feature Indicative Score (NFIS) and plot those scores to demonstrate
the above statement.

\textbf{Normalized Feature Indicative Score:} We use feature indicative score used in the field of text feature selection. Of various feature selection techniques, filtering approach quantifies the relationship between classes and unique features (usually n-gram terms) and selects K number of best features to be fed into a classifier by those metrics. In this experiment, we use Chi-squared text feature selection method \cite{Yang97acomparative}\cite{Zheng04} to obtain feature indicative score for unique terms in our vocabulary for each class. The formulation is as follows.

Let $V$ be our vocabulary and $t \in V$, where t is unique term in $V$. Similarly, let $C$ be a set of unique class labels and $c \in C$, where $|C| = M$. Then, we define $\chi^2$ score for each unique pair of $(t, c)$ 

\begin{equation}
    \chi^2(t,c) = \frac{N*\{p(t,c) p(\bar{t},\bar{c}) - p(t,\bar{c})p(\bar{t},c)\}^2}{p(t)p(\bar{t})p(c)p(\bar{c})}
\end{equation}

, where N is the number of instances in the dataset, $p(t,c)$ is the number of documents containing term t and labeled c over N, $p(t, \bar{c})$ is the number of documents containing term t but not labeled c over N and vice versa.

Next, for each t, we obtain a feature indicative vector, $f_{t}$, in which elements are corresponding to $\chi^2$ between $c$ and $t$, i.e. $f_{t} = \{\chi^2(t,1), \dots, \chi^2(t,M)\}$ and $f_{c} = \{\chi^2(t_{1}, c), \dots, \chi^2(t_{|V|}, c)\}$.

 If $f'_{c} \subset f_{c}$ is the set of $K$ highest $\chi^2$ scores in $f_{c}$, then we can compute  Normalized Feature Indicative Score for $c$ as the sum of all elements, $\chi^2(t_i,c) \in f'_{c}$, and dividing the sum by $argmax(f_{c})$ as a normalizing factor as shown below.
 
\begin{equation}
     NFIS(K;c) = \frac{1}{argmax(f_{c})}\sum_{\chi^2(t_{i}, c) \in f'_{c}} \chi^2(t_{i},c)
  \label{eqn:nfis}
\end{equation}

\begin{figure}[h]
\vspace{-.2in}
  \centering
  \includegraphics[width=0.6\linewidth]{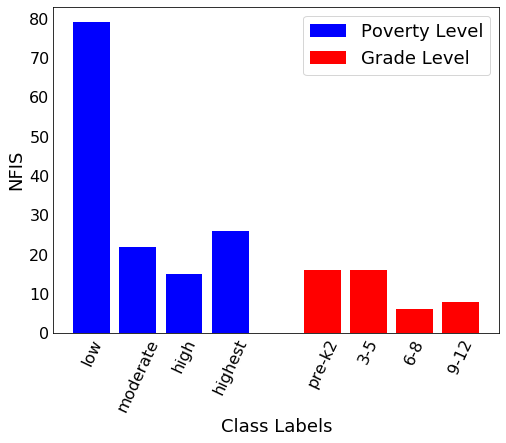}
  \caption{Normalized Feature Indicative Score with K=100 for each class label in poverty level and grade level}
  \label{fig:nfis_dist}
\vspace{-.45in}
\end{figure}

Figure \ref{fig:nfis_dist} shows the distribution of Normalized Feature Indicative Score (NFIS) of poverty level and grade level respectively. The NFIS distribution of poverty level is more imbalanced than the grade level. This indicates that instances (project descriptions) which belong to moderate, high or highest poverty levels are inconsistent, while instances which belong to low poverty level are highly consistent. Having a large variation in the NFIS among a class suggests that instances of certain class labels in the class share a number of indicative features while others don't. In other words, project descriptions did not contain terms that would indicate inclusion in the poverty level labels that the user chose for the project, except for low poverty level. We assume the reason is that the users who chose low poverty level belong to one of groups which are similar to each other, which appears on terms they used in their descriptions. On the contrary, the users who chose other poverty levels are from different populations and did not use common terms that would indicate poverty levels they chose. It is likely that they belong to lower poverty level and didn't describe their poverty levels in the descriptions. The NFIS distribution of grade level supports this argument as instances within the same class are consistent with each other. Since project organizers choose grade levels that accurately reflect their status, each class label has indicative terms that separate themselves from other grade levels. Hence, the NFIS distribution of grade level is more balanced than the NFIS distribution of poverty level.

\subsection{Results}
In this section, we have performed three experiments to understand whether we can trust objective class as supervised class labels and how to detect a subjective class without obtaining reliable labels. Section 4.1 shows that a classifier trained with a objective class shows decent performance measured by f1-macro score, whereas one trained with a subjective class shows significantly lower f1-macro score. However, this result is not sufficient to decide whether a  user-defined class label is subjective since the performance is better than `random guess’ as if a subjective class was a mixture of reliable labels and mislabels. To further examine the distinct characters, we plotted data points in Doc2Vec form compressed by t-SNE method. This plot visually shows that instances are inconsistent in subjective class. Finally, we assumed that features in some class element
are consistent and some are not. As Figure \ref{fig:nfis_dist} shows, our NFIS metric successfully captures the degree of feature inconsistency in each class label as  a subjective class (poverty level) shows highly imbalanced NFIS distribution while an objective class (grade level) shows less imbalance. We summarize our findings as follows:

\vspace{-.05in}
\begin{itemize}
\item Considerably lower f1-macro score for subjective class.
\item Class inconsistency  in subjective class from Doc2Vec method and t-SNE plot.
\item Low feature consistency in subjective class labels from the NFIS scores.
\end{itemize}
\vspace{-.2in}

\section{Subjective Class Analysis on Different Datasets}

In this section, we test the methods used in the previous
section, i.e. 1) classification performance, 2) plotting data points by Doc2Vec and t-SNE, and 3) Normalized Feature Indicative Score on a different type of subjective class i.e., product and service ratings. Customer rating prediction
 \cite{Pang05} is a well-known sentiment analysis
problem that trains a text classifier that takes as input a customer’s review and predicts a rating in a N-star scale.  Pang and Lee \cite{Pang05} found that people are better at discerning the larger difference between ratings than small differences (e.g. a review with one star would be a lot different from a five-star review). This observation allows us to assume that reviews with with 2 - 4 stars would be more ambiguous and less consistent with each other than reviews with 1 or 5 stars. Intuitively, customers who choose polarized values would provide strong indicative features (words) in their reviews, whereas customers who choose 2 - 4 starts might give more neutral reviews. With this hypothesis, we tested framework's ability to detect the subjectivity of customer ratings.

\subsection{User-given Rating Datasets}

We use the Amazon Fine Food Reviews (568,454 reviews) \footnote{\url{https://www.kaggle.com/snap/amazon-fine-food-reviews}} \cite{McAuleyL13} and Yelp   (6,495,456 reviews) \footnote{\url{https://www.yelp.com/dataset}} in our experiments. Both  datasets have ratings in 1 to 5 scale with their reviews given by their users. 

\begin{figure}[ht]
  \begin{subfigure}[b]{0.48\textwidth}
    \caption{Amazon Food Review}
    \includegraphics[width=\textwidth]{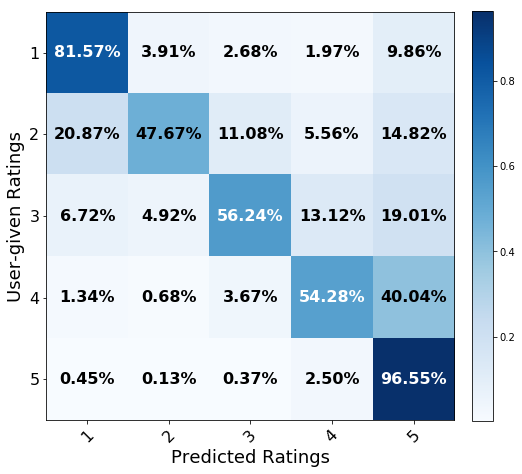}
    \label{fig:f1}
  \end{subfigure}
  \hfill
  \begin{subfigure}[b]{0.48\textwidth}
    \caption{Yelp Review}
    \includegraphics[width=\textwidth]{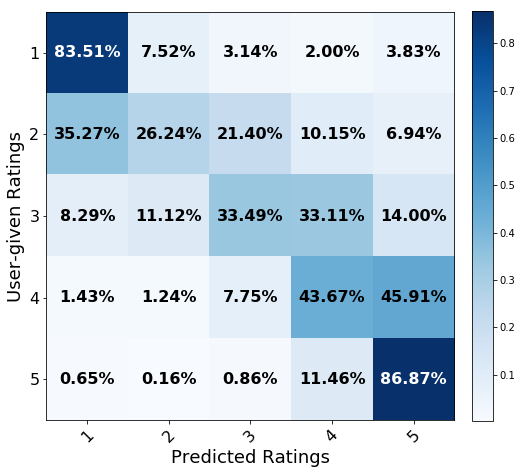}
    \label{fig:f2}
  \end{subfigure}
  \caption{Confusion Matrix of Classifiers for Amazon Food Review and Yelp Review}
  \label{fig:cmat_ratings}
  \vspace{-.2in}
\end{figure}
\subsection{Experiment Description}

\begin{figure}[ht]
  \begin{subfigure}[b]{0.48\textwidth}
    \caption{Amazon Food Review}
    \includegraphics[width=\textwidth]{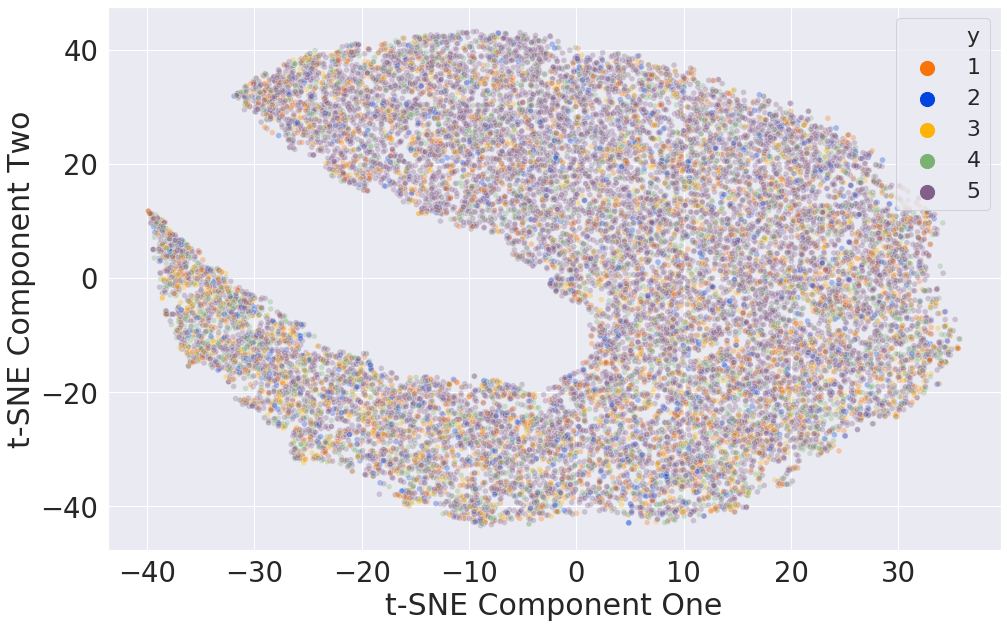}
    \label{fig:f1}
  \end{subfigure}
  \hfill
  \begin{subfigure}[b]{0.48\textwidth}
    \caption{Yelp Review}
    \includegraphics[width=\textwidth]{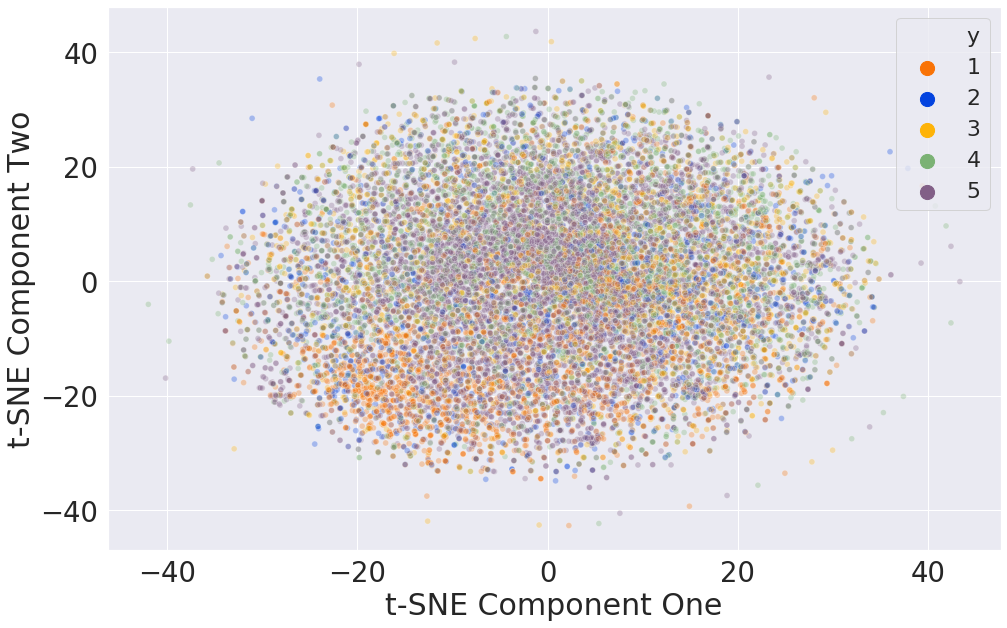}
    \label{fig:f2}
  \end{subfigure}
  \caption{Data Points in Amazon Food Review and Yelp Review by t-SNE}
  \label{fig:tsne_ratings}
\end{figure}

\begin{figure}[h]
  \centering
  \includegraphics[width=0.6\linewidth]{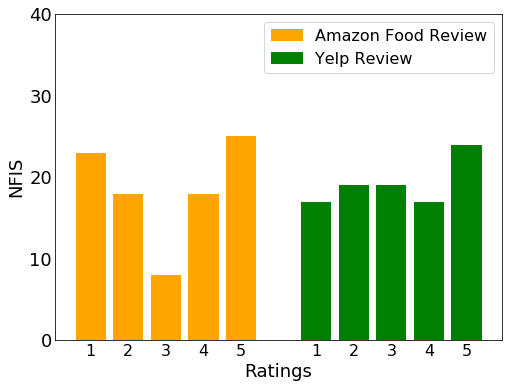}
  \caption{Nomarlized Feature Indicative Score for four different rating datasets}
  \label{fig:nfis_rating}
  \vspace{-.2in}
\end{figure}

We train a linear SVM classifier with penalty rate of 5 for each dataset, and vectorize reviews by Tf-idf bigram BOW features. The distribution of word length for reviews in both dataset represent fat-tail distributions with their peaks around 200 words per review. Thus, we calculate z-score for each instance in each dataset and use instances with z-score less than 1.28 for the training. For the sake of computation, we randomly sample 300,000 instances from selected instances in Yelp dataset. We compute the f1-macro score (equation \ref{eqn:f1score} and \ref{eqn:f1macroscore}) and plot the confusion matrix of each dataset (Figure \ref{fig:cmat_ratings}). For the training of Doc2Vec, we combine both datasets as training corpus as it’s known that the bigger training corpus size yields the better data compression. As a result, we feed 7,063,910 corpora into our PV\_DM model and train it with vector size
= 300, window size = 8, minimum word count = 10, epoch = 10, and 20 iterations for document inference phase. For t-SNE, we train our model with learning rate = 100, iteration = 1000, and perplexity = 20 for Yelp dataset and perplexity = 50 for Amazon dataset (Figure \ref{fig:tsne_ratings}). We choose K=100 to compute Normalized Feature Indicative Score (NFIS) for each rating value (Figure \ref{fig:nfis_rating}), where each marker indicates the same rating value.

\subsection{Subjectivity in Customer Ratings}

As Figure \ref{fig:cmat_ratings} shows each classifier yields high f1 score for rating 1 or 5. The f1-macro score for Amazon dataset is 0.71 and one for Yelp Dataset is 0.55. Also, it’s worth noting that most mis-classifications occur in ratings close to each other in both datasets, e.g. 36\% of reviews labeled rating 2 are misclassified as rating 1, 33\% of reviews labeled rating 3 are misclassified as rating 4. This matches Pang and Lee’s observation  \cite{Pang05} that close ratings are hard to discern. 

From the confusion matrices of two classifiers, we obtain the insight  that customers who give 4-star ratings use the similar positive words used in 5-star ratings, which is why both classifiers predict more than 40\% of 4-star ratings as 5-star ratings in both Yelp Review and Amazon Food Review datasets. On the contrary, customers who give 2 star ratings don't explicitly express their opinions in their reviews, as 39\% of reviews are predicted as higher ratings based purely on their reviews in Yelp Review dataset. As confirmed in Section 4, t-SNE plots for each dataset (Figure \ref{fig:tsne_ratings}) support this insight. Most class labels are mixed with others. In Figure \ref{fig:tsne_ratings} (b), one can observe a cluster for rating 1 on the lower left corner of the cloud, while other ratings hardly show any visible clusters. This shows class inconsistency among those two datasets.

The Normalized Feature Indicative Score (NFIS) (Figure \ref{fig:nfis_rating}) shows that, for both datasets, rating 5 has the strongest score while other ratings show lower scores. This supports that reviews with rating 5 have indicative features while others are less indicative. NFIS score for rating 5 is not as apart from others while the poverty level class labels were quite different from each other. This shows that,
the degree of feature inconsistency in customer ratings is lower than the feature inconsistency in poverty level class, where more than 95\% of labels are inconsistent.

\section{Conclusion and Future Work}
We provide a three step framework for detecting subjective class problem in a dataset. We compare classification performance by f1-macro score and plotting data points by Doc2Vec and t-SNE. We also propose a Normalize Feature Inconsistency Score (NFIS) computed by Chi-square method to effectively understand the inconsistency among labels that have been assigned by users to datasets. Our framework is effective and provides a method of detecting the subjective class problem.  As observed in Section 5, consumer rating systems consists of ambiguous ratings which our framework effectively discerned.  

\vspace{-.05in}
\section*{Acknowledgements}
\vspace{-.1in}
Thanks to Daniel D. Lopez for helping us edit this article.
\vspace{-.05in}

%
%
\bibliographystyle{splncs04}
\bibliography{mybib}

\end{document}